%% file: main.tex
\documentclass[10pt,twocolumn,letterpaper]{article}

 \usepackage[pagenumbers]{cvpr} %

\input{preamble}

\definecolor{cvprblue}{rgb}{0.21,0.49,0.74}
\usepackage[pagebackref,breaklinks,colorlinks,citecolor=cvprblue]{hyperref}

\usepackage{booktabs}
\usepackage{multirow}
\usepackage{multicol}
\usepackage{color, colortbl}
\usepackage{pifont}%
\usepackage{array}

\usepackage{savesym}
\savesymbol{todo}
\usepackage{todonotes}
\restoresymbol{t}{todo}

\title{OccWorld: Learning a 3D Occupancy World Model for Autonomous Driving}

\author{Wenzhao Zheng$^{1,}$\footnotemark[1]\quad Weiliang Chen$^{2,}$\footnotemark[1] \quad Yuanhui Huang$^{1}$ \quad Borui Zhang$^{1}$\quad Yueqi Duan$^{2}$\quad Jiwen Lu$^{1}$ \\
Department of Automation, Tsinghua University, China \\
Department of Electronic Engineering, Tsinghua University, China \\
\texttt{\small wenzhao.zheng@outlook.com; \{chen-wl20,huangyh22,zhang-br21\}@mails.tsinghua.edu.cn;} \\
\texttt{\small \{duanyueqi,lujiwen\}@tsinghua.edu.cn}
}

\begin{document}

\twocolumn[{%
\renewcommand\twocolumn[1][]{#1}%
\vspace{-12mm}
\maketitle
\vspace{-9mm}
\begin{center}
    \centering
    \includegraphics[width=\linewidth]{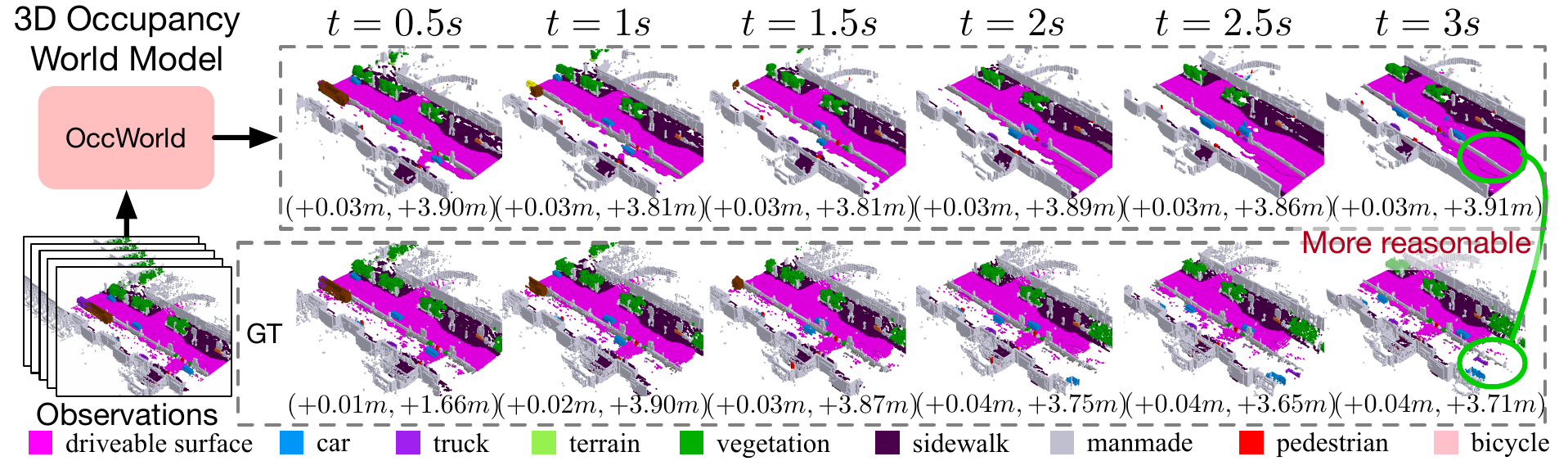}
    \vspace{-7mm}
    \captionof{figure}{
    Given past 3D occupancy observations, our self-supervised OccWorld trained can forecast future scene evolutions and ego movements jointly.
    This task requires a spatial understanding of the 3D scene and temporal modeling of how driving scenarios develop.
    We observe that OccWorld can successfully forecast the movements of surrounding agents and future map elements such as drivable areas.
    OccWorld even generates more reasonable drivable areas than the ground truth, demonstrating its ability to understand the scene rather than memorizing training data.
    Still, it fails to forecast new vehicles entering the sight, which is difficult given their absence in the inputs. 
    }
\label{teaser}
\end{center}%
}]

\renewcommand{\thefootnote}{\fnsymbol{footnote}}
\footnotetext[1]{Equal contribution.}
\renewcommand{\thefootnote}{\arabic{footnote}}

\input{sec/0_abstract}
\input{sec/1_intro}

\input{sec/2_related_work}

\input{sec/3_proposed_approach}

\input{sec/4_experiments}

\input{sec/5_conclusion}

{
    \small
    \bibliographystyle{ieeenat_fullname}
    \bibliography{main}
}

\end{document}

%% file: preamble.tex
\usepackage[dvipsnames]{xcolor}

%% file: sec/0_abstract.tex
\begin{abstract}
Understanding how the 3D scene evolves is vital for making decisions in autonomous driving.
Most existing methods achieve this by predicting the movements of object boxes, which cannot capture more fine-grained scene information.
In this paper, we explore a new framework of learning a world model, OccWorld, in the 3D Occupancy space to simultaneously predict the movement of the ego car and the evolution of the surrounding scenes.
We propose to learn a world model based on 3D occupancy rather than 3D bounding boxes and segmentation maps for three reasons: 
1) \textbf{expressiveness}. 3D occupancy can describe the more fine-grained 3D structure of the scene;
2) \textbf{efficiency}. 3D occupancy is more economical to obtain (e.g., from sparse LiDAR points).
3) \textbf{versatility}. 3D occupancy can adapt to both vision and LiDAR.
To facilitate the modeling of the world evolution, we learn a reconstruction-based scene tokenizer on the 3D occupancy to obtain discrete scene tokens to describe the surrounding scenes.
We then adopt a GPT-like spatial-temporal generative transformer to generate subsequent scene and ego tokens to decode the future occupancy and ego trajectory.
Extensive experiments on the widely used nuScenes benchmark demonstrate the ability of OccWorld to effectively model the evolution of the driving scenes.
OccWorld also produces competitive planning results without using instance and map supervision.
Code: \url{https://github.com/wzzheng/OccWorld}.
\end{abstract}
 

%% file: sec/1_intro.tex
\section{Introduction}
\label{sec:intro}
Autonomous driving has been widely explored in recent years and demonstrated promising results in various scenarios~\cite{tpvformer,cylinder3D,beverse,surroundocc}.
While LiDAR-based models typically show strong performance and robustness in 3D perception due to its capture of structural information~\cite{amvnet,spvnas,af2s3net,drinet++,ye2022lidarmultinet}, 
the more hardware-economical vision-centric solutions have dramatically caught up with the increased perception ability of deep networks~\cite{caddn,bevdepth,lss,bevdet,bevformer}.

Forecasting future scene evolutions is important to the safety of autonomous driving vehicles.
Most existing methods follow a conventional pipeline of perception, prediction, and planning~\cite{hu2023uniad,hu2022stp3,vad}.
Perception aims to obtain a semantic understanding of the surrounding scene such as 3D object detection~\cite{bevdepth,bevdet,bevformer} and semantic map construction~\cite{liao2022maptr,liu2022vectormapnet,li2022hdmapnet,zhang2022beverse}.
The subsequent prediction module captures the motion of other traffic participants~\cite{hu2021fiery, zhang2022beverse, gu2022vip3d, jiang2022pip}, and the planning module then makes decisions based on previous outputs~\cite{renz2022plant,hu2023uniad,hu2022stp3,vad}. 
However, this serial design usually requires ground-truth labels at each stage of training, yet the instance-level bounding boxes and high-definition maps are difficult to annotate.
Furthermore, they usually only predict the motion of object bounding boxes, failing to capture more fine-grained information about the 3D scene.
 
In this paper, we explore a new paradigm to simultaneously predict the evolution of the surrounding scene and plan the future trajectory of the self-driving vehicle.
We propose OccWorld, a world model in the 3D semantic occupancy space, to model the development of the driving scenes. 
We adopt 3D semantic occupancy as the scene representation over the conventional 3D bounding boxes and segmentation maps, which can describe the more fine-grained 3D structure of the scene.
Moreover, 3D occupancy can be effectively learned from sparse LiDAR points~\cite{tpvformer}, and thus is a potentially more economical way to describe the surrounding scenes.
Given the 3D semantic occupancy representation of the current scene, OccWorld aims to predict how it evolves as the self-driving vehicle advances.
To achieve this, we first employ a vector-quantized variational autoencoder (VQVAE)~\cite{oord2017neural} to refine high-level concepts and obtain discrete scene tokens in a self-supervised manner. 
We then tailor the generative pre-training transformers (GPT)~\cite{brown2020language} architecture and propose a spatial-temporal generative transformer to predict the subsequent scene tokens and ego tokens to forecast the future occupancy and ego trajectory, respectively.
We first perform spatial mixing to aggregate scene tokens and obtain multi-scale tokens to represent scenes at multiple levels.
We then apply temporal attention to tokens at different levels to predict tokens for the next frame and use a U-net structure to integrate them.
Finally, we use the trained VQVAE decoder to transform scene tokens to the occupancy space and learn a trajectory decoder to obtain ego planning results.

To demonstrate the effectiveness of OccWorld, we formulate a challenging task of 4D occupancy forecasting, which aims to predict the 3D occupancy of the following frames given a few past frames.
Our OccWorld can effectively forecast future evolutions including moving agents and static elements as shown in Figure~\ref{teaser}, and achieves an average IoU of 26.63 and mIoU of 17.13 for 3$s$ future given 2$s$ history,
OccWorld can also produce planning trajectories with an L2 error of 1.16 without using any instance and map annotations. 
Using self-supervised learned 3D occupancy from camera inputs~\cite{selfocc}, our method achieves non-trivial 4D occupancy forecasting and planning results, demonstrating the potential for interpretable end-to-end autonomous driving without additional human-annotated labels.

%% file: sec/2_related_work.tex
\section{Related Work}
\label{sec:formatting}

\textbf{3D Occupancy Prediction:}
3D occupancy prediction aims to predict whether each voxel in the 3D space is occupied and its semantic label if occupied~\cite{tpvformer,surroundocc,tian2023occ3d,tong2023scene,openoccupancy,pointocc}.
Early methods exploited LiDAR as inputs to complete the 3D occupancy of the entire 3D scene~\cite{js3c,aicnet,dsketch,lmscnet}.
Recent methods began to explore the more challenging vision-based 3D occupancy prediction~\cite{monoscene,tpvformer} or applying vision backbones to efficiently perform LiDAR-based 3D occupancy prediction~\cite{pointocc}.
3D occupancy provides more comprehensive descriptions of the surrounding scene and includes both dynamic and static elements~\cite{tpvformer,surroundocc,pointocc}.
It can also be efficiently learned from sparse accumulated multiple LiDAR scans~\cite{surroundocc}, LiDAR~\cite{tpvformer}, or video sequences~\cite{cao2023scenerf}.
However, existing methods only focus on obtaining the 3D semantic occupancy and ignore its temporal evolution, which is vital to the safety of autonomous driving.
In this paper, we explore the task of 4D occupancy forecasting and propose a 3D occupancy world model to achieve this.

\textbf{World Models for Autonomous Driving:}
World models have a long history in control engineering and artificial intelligence~\cite{sutton1991dyna}, which are usually defined as producing the next scene observation given action and past observations~\cite{ha2018world}.
The development of deep neural networks~\cite{simonyan2014very,szegedy2015going,he2016deep} promoted the use of deep generative models~\cite{goodfellow2014generative,kingma2013auto} as world models.
Based on large pre-trained image generative models like StableDiffusion~\cite{rombach2022high}, recent methods~\cite{gao2023magicdrive, wang2023drivedreamer,yang2023bevcontrol,li2023drivingdiffusion,hu2023gaia} can generate realistic driving sequences of diverse scenarios. 
However, they produce future observations in the 2D image space, lacking understanding of the 3D surrounding scene.
Some other methods explore forecasting point clouds using unannotated LiDAR scans~\cite{khurana2022eo,khurana2023point,mersch2022self,weng2021inverting}, which ignore the semantic information and cannot be applied to vision-based or fusion-based autonomous driving.
Considering this, we explore a world model in the 3D occupancy space to more comprehensively model the 3D scene evolution.

\textbf{End-to-End Autonomous Driving:}
The ultimate goal of autonomous driving is to obtain controlling signals based on observations of the surrounding scenes. 
Recent methods follow this concept to output planning results for the ego car given sensor inputs~\cite{hu2022stp3,hu2023uniad,vad,ye2023fusionad,tong2023scene}.
Most of them follow a conventional pipeline of perception~\cite{tpvformer, beverse,surroundocc,bevdepth,bevformer}, prediction~\cite{zhang2022beverse,liu2021mmtrans,hu2021fiery,gu2022vip3d}, and planning~\cite{zhou2021exploring, vitelli2022safetynet,  huang2023gameformer, huang2023dipp}.
They usually first perform BEV perception to extract relevant information (e.g., 3D agent boxes, semantic maps, tracklets) and then exploit them to infer future trajectories of agents and the ego vehicle.
The following methods incorporated more data~\cite{ye2023fusionad} or extracted more intermediate features~\cite{hu2022stp3,hu2023uniad,vad} to provide more information for the planner, which achieved remarkable performance.
Most methods only model object motions and cannot capture the fine-grained structural and semantic information of the surroundings~\cite{hu2021fiery, zhang2022beverse, gu2022vip3d, jiang2022pip,vad}.
Differently, we propose a world model to predict the evolution of both the surrounding dynamic and static elements.

\begin{figure*}[t]
\centering
\includegraphics[width=\textwidth]{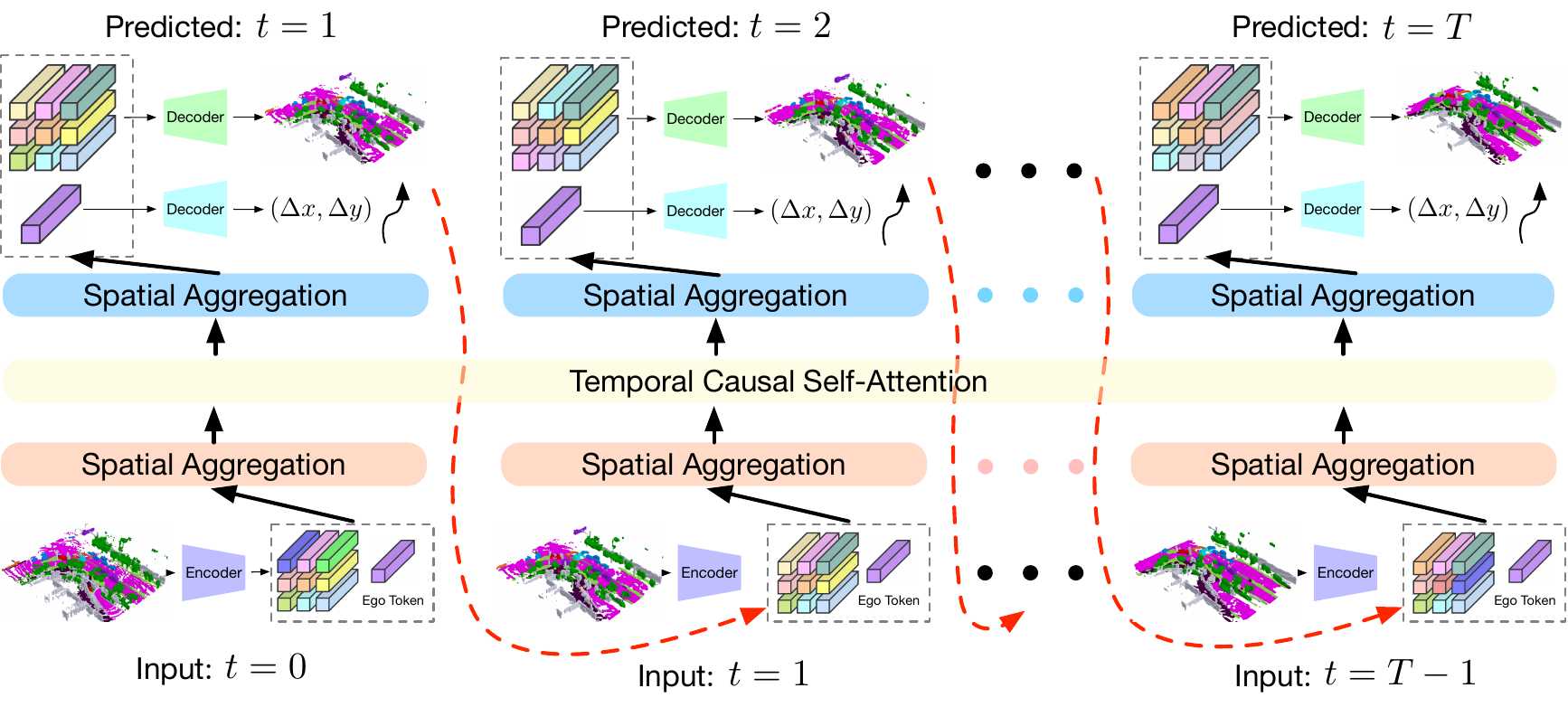}
\vspace{-9mm}
\caption{\textbf{Framework of our OccWorld for 3D semantic occupancy forecast and motion planning.}
We adopt a GPT-like generative architecture to predict the next scene from previous scenes in an autoregressive manner.
We adapt GPT~\cite{brown2020language} to the autonomous driving scenario with two key designs:
1) We train a 3D occupancy scene tokenizer to produce discrete high-level representations of the 3D scene; 
2) We perform spatial mixing before and after spatial-wise temporal causal self-attention to efficiently produce globally consistent scene predictions.
We use ground-truth and predicted scene tokens as inputs for future generations for training and inference, respectively.
}
\label{fig:framework}
\vspace{-5mm}
\end{figure*}

%% file: sec/3_proposed_approach.tex
\section{Proposed Approach}

\subsection{World Model for Autonomous Driving} \label{method sub: 1}
Autonomous driving aims to automatically steer a vehicle to fully prevent or partially reduce actions from human drivers~\cite{hu2023uniad}. 
Formally, the objective of autonomous driving is to obtain the control commands $\mathbf{c}^T$ (e.g., throttle, steer, break) for the present time stamp $T$ given the sensor inputs $\{ \mathbf{s}^{T}, \mathbf{s}^{T-1}, \cdots, \mathbf{s}^{T-t} \}$ from the current and past $t$ frames.

As the mapping from trajectories to control signals is highly dependent on the vehicle specifications and status, the literature usually assumes a given satisfactory controller and thus focuses on trajectory planning for the ego vehicle. 
An autonomous driving model $A$ then takes input as the sensor inputs and ego trajectory from the past $T$ frames and predicts the ego trajectory of future $f$ frames:
\begin{equation}\label{eqn: AD_model}
\begin{aligned}
   & A (\{ \mathbf{s}^{T}, \mathbf{s}^{T-1}, \cdots, \mathbf{s}^{T-t}\}, \{ \mathbf{p}^{T}, \mathbf{p}^{T-1}, \cdots, \mathbf{p}^{T-t}\}) \\
   = & \{ \mathbf{p}^{T+1}, \mathbf{p}^{T+2}, \cdots, \mathbf{p}^{T+f}\},
\end{aligned}
\end{equation}
where $\mathbf{p}^t$ denotes the 3D ego position at the $t$-th time.

The conventional pipeline of autonomous driving usually follows a design of perception, prediction, and planning~\cite{hu2023uniad,hu2022stp3,vad}. 
The perception module $p_{er}$ perceives the surrounding scenes and extracts high-level information $\mathbf{z}$ from the input sensor data $\mathbf{s}$.
The prediction module $p_{re}$ then integrates the high-level information $\mathbf{z}$ to predict the future trajectory $\mathbf{t}_i$ of each agent in the scene.
The planning module $p_{la}$ finally processes the perception and prediction results $\{ \mathbf{z}, \{ \mathbf{t}_i \} \}$ to plan the motion of the ego vehicle.
The conventional pipeline can be formulated as:
\begin{equation}\label{eqn: conventional}
\begin{aligned}
   & p_{la} ( p_{er} (\{ \mathbf{s}^{T}, \cdots, \mathbf{s}^{T-t}\}), p_{re} (p_{er} (\{ \mathbf{s}^{T}, \cdots, \mathbf{}^{T-t}\})))  \\
   = & \{ \mathbf{p}^{T+1}, \mathbf{p}^{T+2}, \cdots, \mathbf{p}^{T+f}\}.
\end{aligned}
\end{equation}

Despite the promising performance of this framework~\cite{hu2023uniad,hu2022stp3,vad}, it usually requires ground-truth labels for supervision at each stage, which can be laborious to annotate.
It only considers object-level movement and fails to model more fine-grained evolutions.

Motivated by this, we explore a new world-model-based autonomous driving paradigm to comprehensively model the evolution of the surrounding scenes and the ego movements.
Inspired by the recent success of generative pre-training transformers (GPT)~\cite{brown2020language} in natural language processing (NLP), we propose an auto-regressive generative modeling framework for autonomous driving scenarios.
We define a world model $w$ to act on scene representations $\mathbf{y}$ and be able to predict future scenes.
Formally, we formulate the function of a world model $w$ as follows:
\begin{equation}\label{eqn: world_model}
\begin{aligned}
    w (\{ \mathbf{y}^{T}, \cdots, \mathbf{y}^{T-t}\}, \{ \mathbf{p}^{T}, \cdots, \mathbf{p}^{T-t}\}) = \mathbf{y}^{T+1}, \mathbf{p}^{T+1}.
\end{aligned}
\end{equation}

Having obtained the predicted scene $\mathbf{y}^{T+1}$ and the ego position $\mathbf{p}^{T+1}$, we can add them to the input and further predict the next frame in an auto-regressive manner, as shown in Figure~\ref{fig:framework}.
The world model $w$ captures the joint distribution of the evolution of the surrounding scene and the ego vehicle, considering their high-order interactions.

\subsection{3D Occupancy Scene Tokenizer} \label{method sub: 2}
As the world model $w$ operates on the scene representation $\mathbf{y}$, its choice is vital to the performance of the world model.
We select $\mathbf{y}$ based on three principles:
1) \textbf{expressiveness}. It should be able to comprehensively contain the 3D structural and semantic information of the 3D scene;
2) \textbf{efficiency}. It should be economical to learn (e.g., from weak supervision or self-supervision);
3) \textbf{versatility}. It should be able to adapt to both vision and LiDAR modalities. 

Considering all the aforementioned principles, we propose to adopt 3D occupancy as the 3D scene representation $\mathbf{y} \in \mathbb{R}^{H \times W \times D}$.
3D occupancy partitions the 3D space surrounding the ego car into $H \times W \times D$ voxels and assigns each voxel with a label $l$ denoting whether it is occupied and which material it is occupied with.
3D occupancy provides a dense representation of the 3D scene and can describe both the 3D structural and semantic information of the scene.
It can be effectively learned from sparse LiDAR annotations~\cite{tpvformer} or potentially from self-supervision of temporal frames~\cite{selfocc}.
3D occupancy is also modality-agnostic and can be obtained from monocular camera~\cite{monoscene}, surrounding cameras~\cite{tpvformer,surroundocc,tong2023scene}, or LiDAR~\cite{pointocc}.

\begin{figure}[t]
\centering
\includegraphics[width=0.475\textwidth]{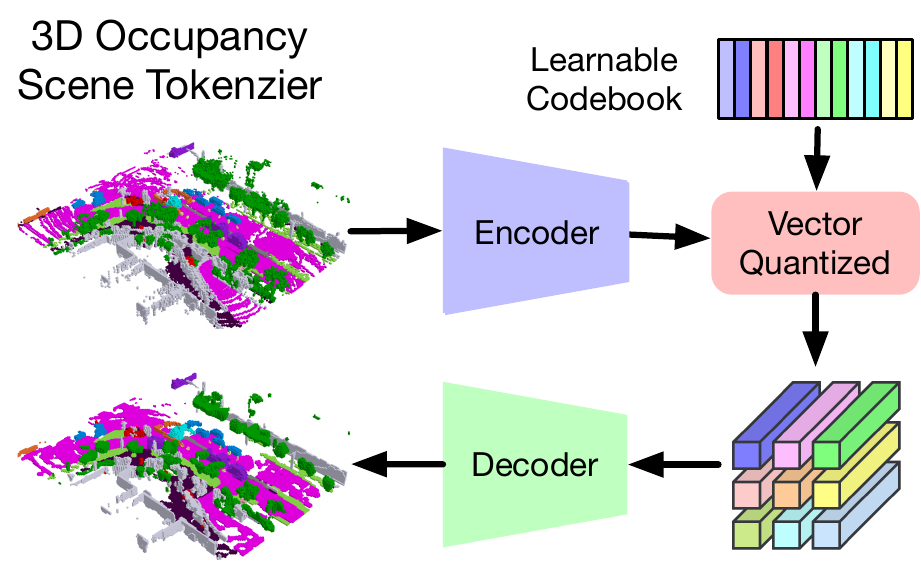}
\vspace{-7mm}
\caption{\textbf{Illustration of the proposed 3D occupancy scene tokenizer.}
We use CNNs to encode the 3D occupancy and perform vector quantization to obtain discrete tokens using a learnable codebook~\cite{oord2017neural}.
We then employ a decoder to reconstruct the input 3D occupancy using the quantized tokens and use a reconstruction objective to train the autoencoder and codebook simultaneously.
}
\label{fig:tokenizer}
\vspace{-6mm}
\end{figure}

Despite its comprehensiveness, 3D occupancy only provides a low-level understanding of the scene, making it difficult to directly model its evolution.
We therefore propose a self-supervised way to tokenize the scene into high-level tokens from 3D occupancy.
We train a vector-quantized autoencoder (VQ-VAE)~\cite{oord2017neural} on $\mathbf{y}$ to obtain discrete tokens $\mathbf{z}$ to better represent the scene, as shown in Figure~\ref{fig:tokenizer}.

For efficiency, we first transform the 3D occupancy $\mathbf{y} \in \mathbb{R}^{H \times W \times D}$ to a BEV representation $\hat{\mathbf{y}} \in \mathbb{R}^{H \times W \times DC'}$ by assigning each category with a learnable class embedding $\in \mathbb{R}^{C'}$ and concatenating them in the height dimension.
We then adopt a lightweight encoder composed of 2D convolution layers to obtain down-sampled features $\hat{\mathbf{z}} \in \mathbb{R}^{\frac{H}{d} \times \frac{W}{d} \times C}$ of the scene, where $d$ is the down-sampling factor.

To obtain a more compact representation, we simultaneously learn a codebook $\mathbf{C} \in \mathbb{R}^{N \times D}$ containing $N$ codes.
Each code $\mathbf{c} \in \mathbb{R}^{C}$ in the codebook encodes a high-level concept of the scene, e.g., whether the corresponding position is occupied by a car.	 
We quantized each spatial feature $\hat{\mathbf{z}}_{ij}$ in $\hat{\mathbf{z}}$ by classifying it to the nearest code $\mathcal{N}(\hat{\mathbf{z}}_{ij}, \mathbf{C})$:
\begin{equation}\label{eqn: quantize}
\begin{aligned}
   \mathbf{z}_{ij} = \mathcal{N}(\hat{\mathbf{z}}_{ij}, \mathbf{C}) = \min_{\mathbf{c} \in \mathbf{C}} ||\hat{\mathbf{z}}_{ij} - \mathbf{c} ||_2,
\end{aligned}
\end{equation}
where $|| \cdot ||_2$ denotes the L2 norm.
We then integrate the quantized features $\{ \mathbf{z}_{ij} \}$ to obtain the final scene representation $\mathbf{z} \in \mathbb{R}^{H \times W \times C}$.

To reconstruct $\widetilde{\mathbf{y}}$ from the learned scene representation $\mathbf{z}$, we use a decoder of 2D deconvolution layers to progressively upsample $\mathbf{z}$ to its original BEV resolution $H \times W \times C''$.
We then perform a split in the channel dimension to reconstruct the height dimension $H \times W \times D \times \frac{C''}{D}$ and apply a softmax layer on each spatial feature to classify them into occupied semantics or unoccupied $H \times W \times D$.

The scene tokenizer transforms 3D occupancy into a more compact discrete space to encode higher-level concepts.
This refined compact space facilitates the modeling of scene evolution for the subsequent world model.

\subsection{Spatial-Temporal Generative Transformer} \label{method sub: 3}

The core of autonomous driving lies in the prediction of how the surrounding world evolves and planning the movement of the ego vehicle accordingly.
While conventional methods usually perform the two tasks separately~\cite{hu2023uniad,hu2022stp3}, we propose to learn a world model $w$ to jointly model the distributions of scene evolution and ego trajectory.

As defined in \eqref{eqn: world_model}, a world model $w$ takes as inputs the past scenes and ego positions and predicts their outcome after driving a certain time interval. 
Based on expressiveness, efficiency, and versatility, we adopt 3D occupancy $\mathbf{y}$ as the scene representation and use a self-supervised tokenizer to obtain high-level scene tokens $\mathbf{T} = \{ \mathbf{z}_i \}$.
To integrate the ego movement, we further aggregate $\mathbf{T}$ with an ego token $\mathbf{z}_0 \in \mathbb{R}^C$ to encode the spatial position of the ego vehicle. 

The proposed OccWorld $w$ then functions on the world tokens $\mathbf{T}$, which can be formulated as:
\begin{equation}\label{eqn: OccWorld}
\begin{aligned}
   w (\mathbf{T}^T, \cdots, \mathbf{T}^{T-t}) = \mathbf{T}^{T+1},
\end{aligned}
\end{equation}
where $T$ is the current time stamp, and $t$ is the number of history frames available. 

Inspired by the remarkable sequential prediction performance of GPT~\cite{brown2020language}, we adopt a GPT-like autoregressive transformer architecture to instantiate \eqref{eqn: OccWorld}.
However, the migration of GPT from natural language processing to the autonomous driving scenario is not trivial.
GPTs predict a single token each time, while the world model $w$ in autonomous driving is required to predict a set of tokens $\mathbf{T}$ as the next future.
Due to the vast number of world tokens, directly leveraging the GPT architecture to predict each token $\in \mathbf{T}^{T+1}$ is both inefficient and ineffective.

\begin{figure}[t]
\centering
\includegraphics[width=0.475\textwidth]{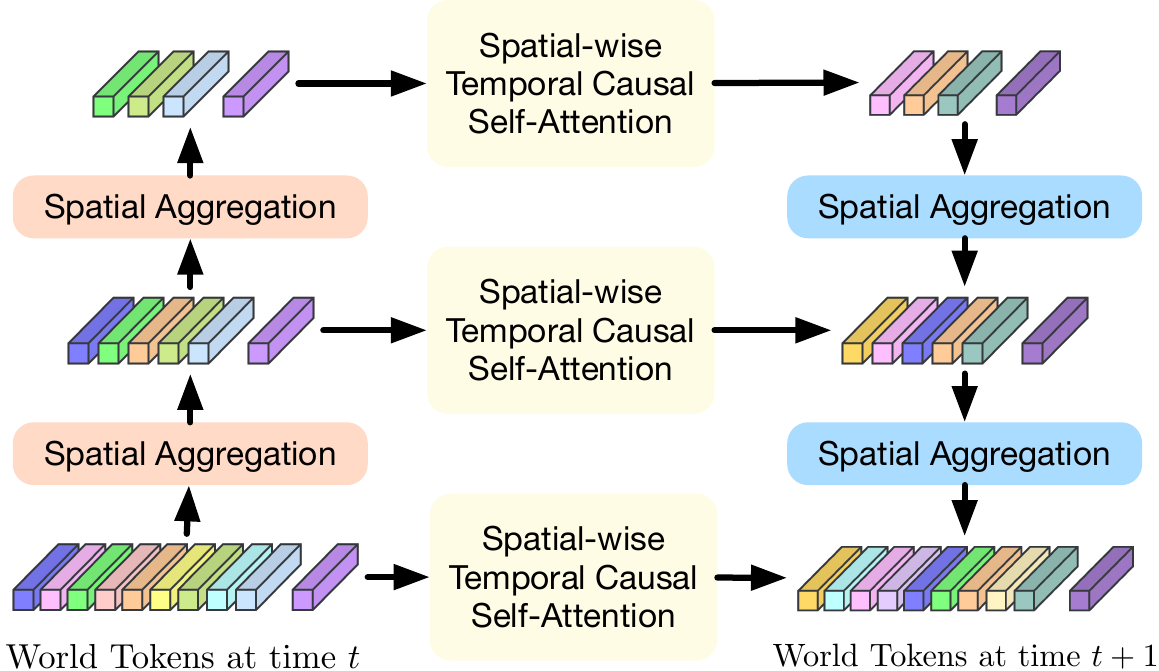}
\vspace{-7mm}
\caption{\textbf{Illustration of the proposed spatial-temporal generative transformer.}
As each scene is composed of numerous world tokens, we adopt spatial mixing modules to model their intrinsic dependencies and obtain multi-scale world tokens to capture multi-level information.
We then perform spatial-wise temporal causal self-attention at each level to forecast the next scene.
We employ a U-net structure to aggregate the multi-scale predictions.
}
\label{fig:transformer}
\vspace{-6mm}
\end{figure}

Both the spatial relations of world tokens within each time stamp and the temporal relations of tokens across different time stamps should be considered to comprehensively model the world evolution.
Therefore, we propose a spatial-temporal generative transformer architecture to effectively process past world tokens and make predictions of the next future, as shown in Figure~\ref{fig:transformer}.

We apply spatial aggregation (e.g., self-attention~\cite{vit}) to world tokens $\mathbf{T}$ to enable interactions between scene tokens as well as ego tokens.
We then merge the scene tokens in each $2 \times 2$ window with a stride of 2 and thus down-sample the scene tokens by a factor of 4.
We repeat this procedure for $K$ times to obtain world tokens of hierarchical scales $\{\mathbf{T}_0, \cdots, \mathbf{T}_K \}$ to describe the 3D scene at different levels.

We use several sub-world models $w = \{w_0, \cdots, w_K \}$ to predict the future at different spatial scales.
For each sub-world model $w_i$, we impose temporal attention on the tokens $\{ \mathbf{z}^{T}_{j,i}, \cdots, \mathbf{z}^{T-t}_{j,i} \}$ at each position $j$ to obtain the predicted corresponding token $\mathbf{z}^{T+1}_{j,i}$ of the next frame: 
\begin{equation}\label{eqn: predict}
\begin{aligned}
  \hat{\mathbf{z}}^{T+1}_{j,i}  = \text{TA} (\mathbf{z}^{T}_{j,i}, \cdots, \mathbf{z}^{T-t}_{j,i}),
\end{aligned}
\end{equation}
where TA denotes masked temporal attention which blocks the effect of future tokens to previous tokens.
$\mathbf{z}^{t}_{j,i} \in \mathbf{T}^{t}_{i}$ represents the $j$-th world token of the $i$-th scale at time stamp $t$.
We finally employ a U-net structure to aggregate predicted tokens at different scales to ensure spatial consistency.

Our spatial-temporal generative transformer can model the world evolution in driving sequences considering the joint distributions of world tokens within each time and across time.
The temporal attention predicts the evolution of a fixed position in the surrounding area, while the spatial aggregation makes each token aware of the global scene.

\subsection{OccWorld: a 3D Occupancy World Model} \label{method sub: 4}
We present the overall training framework of our OccWorld model for autonomous driving. 
Having obtained the forecasted world tokens, we reuse the scene decoder $d$ to decode the predicted 3D occupancy $\hat{\mathbf{y}}^{T+1} = d(\hat{\mathbf{z}}^{T+1})$ and additionally learn an ego decoder $d_{ego}$ to produce the ego displacement $\hat{p}^{T+1} = d_{ego}(\hat{z}^{T+1}_0) $ w.r.t the current frame.

We adopt a two-stage training strategy to effectively train our OccWorld.
For the first stage, we train the scene tokenizer $e$ and decoder $d$ using 3D occupancy loss~\cite{tpvformer}:
\begin{equation}\label{eqn: loss_tokenzier}
\begin{aligned}
  J_{e,d} = L_{soft} (d(e(\mathbf{y})), \mathbf{y}) + \lambda_1 \  L_{lovasz} (d(e(\mathbf{y})), \mathbf{y}),
\end{aligned}
\end{equation}
where $L_{soft}$ and $L_{lovasz}$ is the softmax and lovasz-softmax loss~\cite{lovasz}, respectively, and $\lambda_1$ is a balance factor.

For the second stage, we adopt the learned scene tokenizer $e$ to obtain scene tokens $\mathbf{z}$ for all the frames and constrain the discrepancy between predicted tokens $\hat{\mathbf{z}}$ and $\mathbf{z}$.
We then apply the softmax loss to enforce the correct classification of $\hat{\mathbf{z}}$ to the correct codes in the codebook $\mathbf{C}$ as $\mathbf{z}$.
For the ego token, we simultaneously learn the ego decoder $d_{ego}$ and apply L2 loss on the predicted displacement $\hat{p} = d_{ego}(\hat{\mathbf{z}}_0) $ and the ground-truth one $\mathbf{p}$.
The overall objective for the second stage can be formulated as follows:
\begin{equation}\label{eqn: loss_world}
\begin{aligned}
  J_{w,d_{ego}} = \sum_{t=1}^T (\sum_{j=1}^{M_0}  L_{soft} (\hat{\mathbf{z}}_{j,0}^t , \mathbf{C}(\mathbf{z}_{j,0}^t) \\
  + \  \lambda_2 \  L_{L2} (d_{ego}(\hat{\mathbf{z}}^t_0), \mathbf{p}^t)),
\end{aligned}
\end{equation}
where T and $M_0$ are the numbers of frames and spatial tokens of the original scale, respectively.
$\mathbf{C}(\cdot)$ denotes the index of the corresponding code in the codebook $\mathbf{C}$.
$L_{L2}$ measures the L2 discrepancy between two trajectories.

For efficient training, we use tokens obtained by the scene tokenizer $e$ as inputs but apply masked temporal attention~\cite{brown2020language} to block the effect of future tokens.
During inference, we progressively predict world tokens of the next frame using predicted tokens of past frames.

Our OccWorld can be applied to various types of 3D occupancy to adapt to different settings (e.g., end-to-end autonomous driving).
The scene representation model $r$ can be an oracle providing ground-truth occupancy, or a perception model taking images or LiDAR as inputs.
Different from the conventional perception, predicting, and planning pipeline, OccWorld models the joint evolution of the surrounding scene and the ego movement to capture high-order interactions between the ego vehicle and the environment.
Combined with machine-annotated~\cite{surroundocc}, LiDAR-collected~\cite{tpvformer}, or self-supervised~\cite{selfocc} 3D occupancy, OccWorld has the potential to scale up to large-scale training, paving the way for large driving models.

%% file: sec/4_experiments.tex
\begin{table*}[t]
\setlength{\tabcolsep}{0.007\linewidth}
\caption{\textbf{4D occupancy forecasting performance.}
Aux. Sup. denotes auxiliary supervision apart from the ego trajectory.
Avg. denotes the average performance of that in 1s, 2s, and 3s.
We use bold numbers to denote the best results.
}
\vspace{-4mm}
\centering
\begin{tabular}{l|cc|ccccc|ccccc|c}
\toprule
\multirow{2}{*}{Method} & \multirow{2}{*}{Input} & \multirow{2}{*}{Aux. Sup.} &
\multicolumn{5}{c|}{mIoU (\%) $\uparrow$} & 
\multicolumn{5}{c|}{IoU (\%) $\uparrow$} &
\cellcolor{gray!30} \\
&& & 0s & 1s & 2s & 3s & \cellcolor{gray!30}Avg. & 0s & 1s & 2s & 3s & \cellcolor{gray!30}Avg. & \cellcolor{gray!30}\multirow{-2}*{FPS} \\
\midrule
\textbf{Copy\&Paste} & 3D-Occ & None & 66.38 & 14.91 & 10.54 & 8.52 & \cellcolor{gray!30}11.33 & 62.29 & 24.47 & 19.77 & 17.31 & \cellcolor{gray!30}20.52 & \cellcolor{gray!30}- \\
\textbf{OccWorld-O} & 3D-Occ & None & \textbf{66.38} & \textbf{25.78} & \textbf{15.14} & \textbf{10.51} & \cellcolor{gray!30}\textbf{17.14}  & \textbf{62.29} & \textbf{34.63} & \textbf{25.07} & \textbf{20.18} & \cellcolor{gray!30}\textbf{26.63}  & \cellcolor{gray!30}\textbf{18.0} \\
\hline
\textbf{OccWorld-D} & Camera & 3D-Occ & 18.63 & 11.55 & 8.10 & 6.22 & \cellcolor{gray!30}8.62 & 22.88 & 18.90 & 16.26 & 14.43 & \cellcolor{gray!30}16.53  & \cellcolor{gray!30}2.8 \\
\textbf{OccWorld-T} & Camera & Semantic LiDAR & 7.21 & 4.68 &  3.36 & 2.63 & \cellcolor{gray!30}3.56 & 10.66 & 9.32 & 8.23 & 7.47 & \cellcolor{gray!30}8.34  & \cellcolor{gray!30}2.8 \\
\textbf{OccWorld-S} & Camera & None & 0.27 & 0.28 & 0.26 & 0.24 & \cellcolor{gray!30}0.26 & 4.32 & 5.05 & 5.01 & 4.95 & \cellcolor{gray!30}5.00 & \cellcolor{gray!30}2.8 \\
\bottomrule
\end{tabular}%
\label{forecast}
\vspace{-4mm}
\end{table*}

\begin{figure*}[t]
\centering
\includegraphics[width=\textwidth]{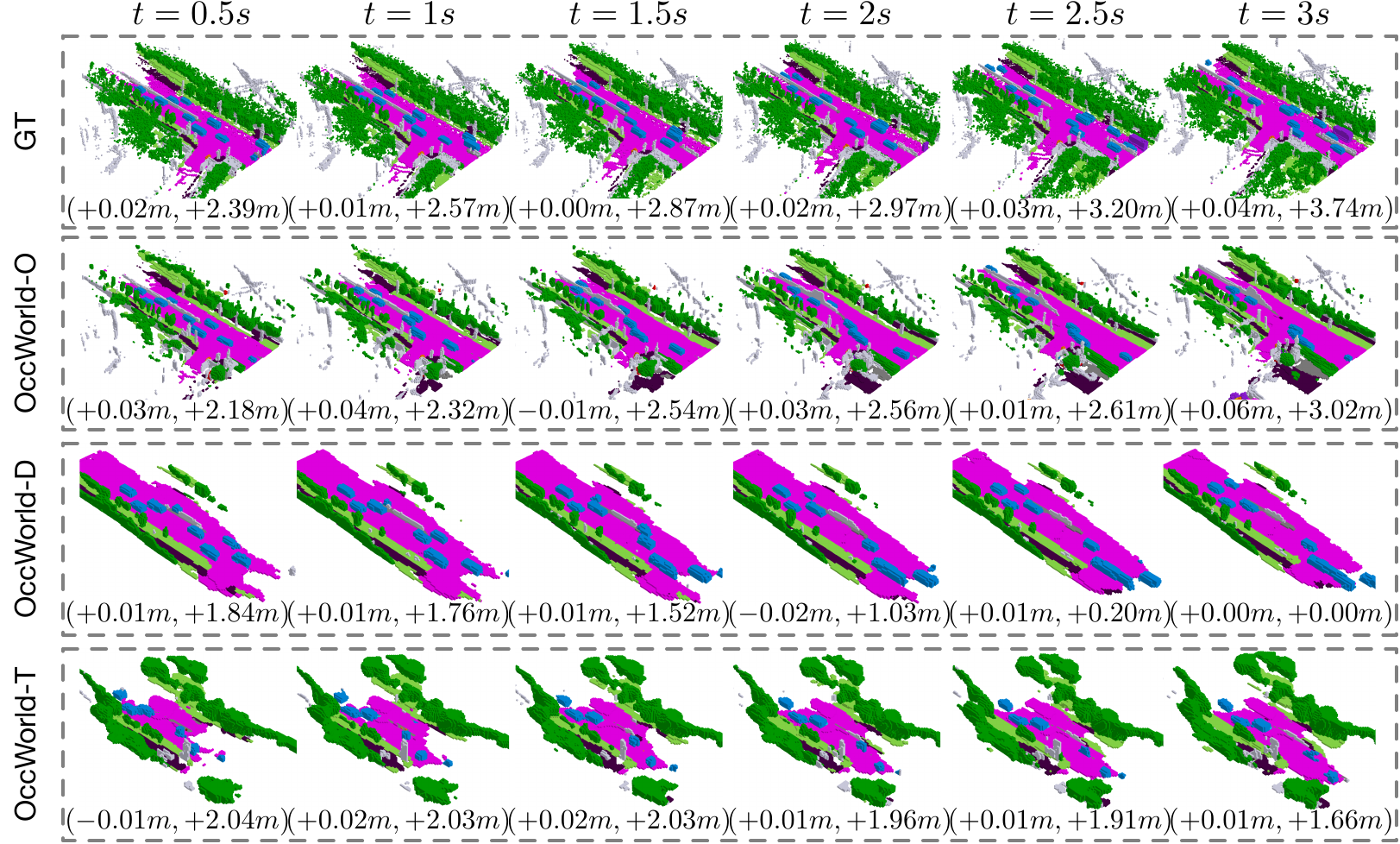}
\vspace{-7mm}
\caption{\textbf{Visualizations of the forecasting and planning results of OccWorld-O, OccWorld-D, and OccWorld-T.}
}
\label{fig:vis}
\vspace{-5mm}
\end{figure*}

\section{Experiments}

\subsection{Task Descriptions}
In this paper, we explore a world-model-based framework for autonomous driving and propose OccWorld to model the joint evolutions of ego trajectory and scene evolutions.
We conduct two tasks to evaluate our OccWorld: 4D occupancy forecasting on the Occ3D dataset~\cite{tian2023occ3d} and motion planning on the nuScenes dataset~\cite{nuscenes}.
We present the dataset and evaluation metric details in the supplementary material.

\textbf{4D occupancy forecasting.}
3D occupancy prediction aims to reconstruct the semantic occupancy for each voxel in the surrounding space, which cannot capture the temporal evolution of the 3D occupancy.
In this paper, we explore the task of 4D occupancy forecasting, which aims to forecast the future 3D occupancy given a few historical occupancy inputs.
We use mIoU and IoU as the evaluation metric.

\begin{table*}[t]
\setlength{\tabcolsep}{0.008\linewidth}
\caption{\textbf{Motion planning performance.} 
Aux. Sup. denotes auxiliary supervision apart from the ego trajectory.
We use bold and underlined numbers to denote the best and second-best results, respectively.
$^\dagger$ denotes using the metric computation adopted in VAD~\cite{vad}.
}
\vspace{-3mm}
\centering
\begin{tabular}{l|cc|cccc|cccc|c}
\toprule
\multirow{2}{*}{Method} & \multirow{2}{*}{Input} & \multirow{2}{*}{Aux. Sup.} &
\multicolumn{4}{c|}{L2 (m) $\downarrow$} & 
\multicolumn{4}{c|}{Collision Rate (\%) $\downarrow$} &
\cellcolor{gray!30} \\
&& & 1s & 2s & 3s & \cellcolor{gray!30}Avg. & 1s & 2s & 3s & \cellcolor{gray!30}Avg. & \cellcolor{gray!30}\multirow{-2}*{FPS} \\
\midrule
IL~\cite{ratliff2006maximum} & LiDAR & None  & \underline{0.44} & 1.15 & 2.47 & \cellcolor{gray!30}1.35 & 0.08 & 0.27 & 1.95 & \cellcolor{gray!30}0.77 & \cellcolor{gray!30}- \\
NMP~\cite{zeng2019nmp} & LiDAR & Box \& Motion & 0.53 & 1.25 & 2.67 & \cellcolor{gray!30}1.48 & \textbf{0.04} & \underline{0.12} & \underline{0.87} & \cellcolor{gray!30}0.34 & \cellcolor{gray!30}- \\
FF~\cite{hu2021ff} & LiDAR & Freespace  & 0.55 & 1.20 & 2.54 & \cellcolor{gray!30}1.43 & 0.06 & 0.17 & 1.07 & \cellcolor{gray!30}0.43 & \cellcolor{gray!30}- \\
EO~\cite{khurana2022eo} & LiDAR & Freespace  & 0.67 & 1.36 & 2.78 & \cellcolor{gray!30}1.60 & \textbf{0.04} & \textbf{0.09} & 0.88 & \cellcolor{gray!30}\underline{0.33} & \cellcolor{gray!30}- \\
\midrule
ST-P3~\cite{hu2022stp3} & Camera & Map \& Box \& Depth & 1.33 & 2.11 & 2.90 & \cellcolor{gray!30}2.11 & 0.23 & 0.62 & 1.27 & \cellcolor{gray!30}0.71 &  \cellcolor{gray!30}1.6 \\
UniAD~\cite{hu2023uniad} & Camera & { \footnotesize Map \& Box \& Motion \& Tracklets \& Occ}  & 0.48 & \textbf{0.96} & \textbf{1.65} & \cellcolor{gray!30}\textbf{1.03} & \underline{0.05} & 0.17 & \textbf{0.71} & \cellcolor{gray!30}\textbf{0.31} & \cellcolor{gray!30}1.8 \\
VAD-Tiny~\cite{vad}  & Camera & Map \& Box \& Motion  & 0.60 & 1.23 & 2.06 & \cellcolor{gray!30}1.30 & 0.31 & 0.53 & 1.33 & \cellcolor{gray!30}0.72 & \cellcolor{gray!30}\underline{16.8} \\
VAD-Base~\cite{vad} & Camera & Map \& Box \& Motion & 0.54 & 1.15 & \underline{1.98} & \cellcolor{gray!30}1.22 & \textbf{0.04} & 0.39 & 1.17 & \cellcolor{gray!30}0.53 & \cellcolor{gray!30}4.5 \\
OccNet~\cite{tong2023scene} & Camera & 3D-Occ \& Map \& Box & 1.29 & 2.13 & 2.99 & \cellcolor{gray!30}2.14 & 0.21 & 0.59 & 1.37 & \cellcolor{gray!30}0.72  & \cellcolor{gray!30}2.6 \\
\midrule
OccNet~\cite{tong2023scene} & 3D-Occ & Map \& Box & 1.29 & 2.31 & 2.98 & \cellcolor{gray!30}2.25 & 0.20 & 0.56 & 1.30 & \cellcolor{gray!30}0.69  & \cellcolor{gray!30}- \\
\textbf{OccWorld-O} & 3D-Occ & None & \textbf{0.43} & \underline{1.08} & 1.99 & \cellcolor{gray!30}\underline{1.17} & 0.07 & 0.38 & 1.35 & \cellcolor{gray!30}0.60  & \cellcolor{gray!30}\textbf{18.0} \\
\midrule
\textbf{OccWorld-D} & Camera & 3D-Occ & 0.52 & 1.27 & 2.41 & \cellcolor{gray!30}1.40 & 0.12 & 0.40 & 2.08 & \cellcolor{gray!30}0.87  & \cellcolor{gray!30}2.8 \\
\textbf{OccWorld-T} & Camera & Semantic LiDAR & 0.54 & 1.36 & 2.66 & \cellcolor{gray!30}1.52 & 0.12 & 0.40 & 1.59 & \cellcolor{gray!30}0.70  & \cellcolor{gray!30}2.8 \\
\textbf{OccWorld-S} & Camera & None & 0.67 & 1.69 & 3.13 & \cellcolor{gray!30}1.83 & 0.19 & 1.28 & 4.59 & \cellcolor{gray!30}2.02 & \cellcolor{gray!30}2.8 \\
\midrule
\midrule
\color{gray}VAD-Tiny$^\dagger$~\cite{vad} & \color{gray} Camera & \color{gray} Map \& Box \& Motion  & \color{gray} 0.46 & \color{gray} 0.76 & \color{gray} 1.12 & \color{gray} \cellcolor{gray!30}0.78 & \color{gray} 0.21 & \color{gray}0.35 & \color{gray}0.58 & \color{gray}\cellcolor{gray!30}0.38 & \color{gray}\cellcolor{gray!30}\underline{16.8} \\
\color{gray}VAD-Base$^\dagger$~\cite{vad} & \color{gray}Camera & \color{gray}Map \& Box \& Motion & \color{gray} {0.41} & \color{gray}\underline{0.70} & \color{gray}\underline{1.05} & \color{gray}\cellcolor{gray!30}\underline{0.72} & \color{gray}\underline{0.07} & \color{gray}\textbf{0.17} & \color{gray}\textbf{0.41} & \color{gray}\cellcolor{gray!30}\textbf{0.22} & \color{gray}\cellcolor{gray!30}4.5 \\
\midrule
\color{gray}\textbf{OccWorld-O}$^\dagger$ & \color{gray}3D-Occ & \color{gray}None & \color{gray}\textbf{0.32} & \color{gray}\textbf{0.61} &\color{gray}\textbf{0.98} & \color{gray}\cellcolor{gray!30}\textbf{0.64} & \color{gray}\textbf{0.06} & \color{gray} 0.21 & \color{gray}\underline{0.47} & \color{gray}\cellcolor{gray!30}\underline{0.24}  & \color{gray}\cellcolor{gray!30}\textbf{18.0} \\
\color{gray}\textbf{OccWorld-D}$^\dagger$ & \color{gray}Camera & \color{gray}3D-Occ & \color{gray}\underline{0.39} & \color{gray}0.73 & \color{gray}1.18 & \color{gray}\cellcolor{gray!30} 0.77 & \color{gray}0.11 & \color{gray}\underline{0.19} & \color{gray}0.67 & \color{gray}\cellcolor{gray!30} 0.32  & \color{gray}\cellcolor{gray!30}2.8 \\
\color{gray}\textbf{OccWorld-T}$^\dagger$ & \color{gray}Camera & \color{gray}Semantic LiDAR & \color{gray}0.40 & \color{gray}0.77 & \color{gray}1.28 & \color{gray}\cellcolor{gray!30} 0.82 & \color{gray}0.12 & \color{gray}0.22 & \color{gray}0.56 & \color{gray}\cellcolor{gray!30} 0.30 & \color{gray}\cellcolor{gray!30}2.8 \\
\color{gray}\textbf{OccWorld-S}$^\dagger$ & \color{gray} Camera & \color{gray} None & \color{gray} 0.49 & \color{gray} 0.95 & \color{gray} 1.55 & \color{gray} \cellcolor{gray!30}0.99 & \color{gray} 0.19& \color{gray} \color{gray} 0.56 & \color{gray} 1.54 &  \color{gray} \cellcolor{gray!30} 0.76 & \color{gray} \cellcolor{gray!30}2.8 \\
\bottomrule
\end{tabular}%
\label{tab:sota-plan}
\vspace{-7mm}
\end{table*}

\textbf{Motion planning.}
The objective of motion planning is to produce safe future trajectories for the self-driving vehicle given ground-truth surrounding information or perception results.
The planned trajectory is represented by a series of 2D waypoints in the BEV plane (ground plane).
We use L2 error and collision rate as the evaluation metric.

\subsection{Implementation Details}
We followed existing works~\cite{vad,hu2023uniad} and used a 2-second historical context to forecast the subsequent 3 seconds.
 The scene tokenizer employs a down-sampling factor of 4, featuring a codebook comprising 512 nodes and a 128-dimensional feature representation.
The spatial-temporal generative transformer comprises 3 scales, each incorporating 6 layers of spatial-wise temporal attention for scene tokens with 2 layers of spatial cross-attention and temporal cross-attention for ego planning tokens.

During training, we applied mask operations to all temporal attention mechanisms to prevent the influence of future information on forecasting.
For inference, we employ autoregressive prediction to foresee 3 seconds into the future based on a 2-second historical context.
We adopted the AdamW optimizer~\cite{loshchilov2017adamw} and a Cosine Annealing scheduler~\cite{loshchilov2016cosineanneal} for training.
We set an initial learning rate of 1 × $10^{-3}$ and the weight decay at 0.01 and.
We use a batch size of 1 per GPU on 8 NVIDIA GeForce RTX 4090 GPUs.

\begin{table*}[t]
\setlength{\tabcolsep}{0.016\linewidth}
\caption{\textbf{Effect of different hyperparameters for the scene tokenizer.} 
We use bold numbers to denote the best results.
}
\vspace{-4mm}
\centering
\begin{tabular}{c|cc|cccc|cccc|c}
\toprule
 \multirow{2}{*}{Setting} & \multicolumn{2}{c|}{Reconstruction}  &
\multicolumn{4}{c|}{Forecasting mIoU (\%) $\uparrow$} & 
\multicolumn{4}{c|}{Planning L2 (m) $\downarrow$} &
\cellcolor{gray!30} \\
& mIoU $\uparrow$ & IoU $\uparrow$ & 1s & 2s & 3s & \cellcolor{gray!30}Avg. & 1s & 2s & 3s & \cellcolor{gray!30}Avg. & \cellcolor{gray!30}\multirow{-2}*{FPS} \\
\midrule
($50^2$, 128, 512) & 66.38 & 62.29 & \textbf{25.78} & \textbf{15.14} & 10.51 & \cellcolor{gray!30}\textbf{17.14} & 0.43 & \textbf{1.08} & 1.99 & \cellcolor{gray!30}1.17  & \cellcolor{gray!30}18.0 \\
($50^2$, 128, 256) & 63.40 & 60.33 & 24.25 & 14.34 & 10.13 & \cellcolor{gray!30}16.24 & \textbf{0.42} & \textbf{1.08} & \textbf{1.95} & \cellcolor{gray!30}\textbf{1.15}  & \cellcolor{gray!30}17.8 \\
 ($50^2$, 128, 1024) & 60.50 & 59.07 & 23.55 & 14.66 & \textbf{10.68} & \cellcolor{gray!30}16.30 & 0.47 & 1.18 & 2.19 & \cellcolor{gray!30}1.28  & \cellcolor{gray!30}17.8 \\
 ($25^2$, 256, 512) & 36.28 & 44.02 & 12.10 & 8.13 & 6.20 & \cellcolor{gray!30}8.81 & 3.27 & 6.54 & 9.78 & \cellcolor{gray!30}6.53  & \cellcolor{gray!30}\textbf{28.1} \\
 ($100^2$, 128, 512) & \textbf{78.12} & \textbf{71.63} & 18.71 & 10.75 & 7.68 & \cellcolor{gray!30}12.38 & 0.50 & 1.25 & 2.33 & \cellcolor{gray!30}1.36  & \cellcolor{gray!30}6.7 \\
 ($50^2$, 64, 512) & 64.98 & 61.50 & 21.83 & 12.90 & 9.28 & \cellcolor{gray!30}14.67 & 0.49 & 1.24 & 2.26 & \cellcolor{gray!30}1.33  & \cellcolor{gray!30}20.1 \\
\bottomrule
\end{tabular}%
\label{tab:vae}
\vspace{-7mm}
\end{table*}

\begin{table}[t]
\setlength{\tabcolsep}{0.02\linewidth}
\caption{\textbf{Ablation study of the spatial-temporal generative transformer.} 
We report average results over the 1s, 2s, and 3s.
}
\vspace{-4mm}
\centering
\begin{tabular}{l|cc|cc|c}
\toprule
\multirow{2}{*}{Method}& \multicolumn{2}{c|}{Forecast}  &
\multicolumn{2}{c|}{Planning} & 
\cellcolor{gray!30} \\
& mIoU$\uparrow$ & IoU$\uparrow$ & L2$\downarrow$ & Col.$\downarrow$ & \cellcolor{gray!30}\multirow{-2}*{FPS} \\
\midrule
OccWorld-O & \textbf{17.14} & \textbf{26.63} & \textbf{1.17} & \textbf{0.60} & \cellcolor{gray!30}18.0 \\
w/o spatial attn & 10.07 & 21.44 & 1.42 & 1.21 & \cellcolor{gray!30}\textbf{28.6} \\
w/o temporal attn & 8.98 & 20.10 & 2.06 & 2.56 & \cellcolor{gray!30}26.5 \\
w/o ego & 15.13 & 24.66 & - & - & \cellcolor{gray!30}18.8 \\
w/o ego temporal & 12.07 & 23.09 & 5.89 & 6.23 & \cellcolor{gray!30}18.5 \\
\bottomrule
\end{tabular}%
\label{tab:world}
\vspace{-7mm}
\end{table}

\subsection{Results and Analysis}

\textbf{4D occupancy forecasting.}
We evaluated the 4D occupancy forecasting performance of our OccWorld in several settings: OccWorld-O (using ground-truth 3D occupancy), OccWorld-D (using predicted results of TPVFormer~\cite{tpvformer} trained with dense ground-truth 3D occupancy), OccWorld-T (using predicted results of TPVFormer~\cite{tpvformer} trained with sparse semantic LiDAR\footnote{\url{https://github.com/wzzheng/TPVFormer}}), and OccWorld-S (using predicted results of TPVFormer~\cite{selfocc} trained in a self-supervised manner\footnote{\url{https://github.com/huang-yh/SelfOcc}}).
Copy\&Paste denotes copying the current ground-truth occupancy as future observations.
The 0s results represent the reconstruction accuracy.

We compare the performance of the aforementioned settings in Table~\ref{forecast}.
We observe that OccWorld-O can generate non-trivial future 3D occupancy with much better results than Copy\&Paste, showing that our model learns the underlying scene evolution.
OccWorld-D, OccWorld-T, and OccWorld-S can be seen as end-to-end vision-based 4D occupancy forecasting methods as they take surrounding images as input.
This task is very challenging since it requires both 3D structure reconstruction and forecasting.
It is especially difficult for the self-supervised OccWorld-S, which exploits no 3D occupancy information even during training.
Still, our OccWorld generates future 3D occupancy with non-trivial mIoU and IoU on the end-to-end setting.

\textbf{Visualizations.}
We visualize the output results of the proposed OccWorld in Figure~\ref{fig:vis}.
We see that our models can successfully forecast the movements of cars and can complete unseen map elements in the inputs such as drivable areas.
The planning trajectory is also more accurate with better 4D occupancy forecasting.

\textbf{Motion planning.}
We compare the motion planning performance of the proposed OccWorld with state-of-the-art end-to-end autonomous driving methods, as shown in Table~\ref{tab:sota-plan}.
We also evaluate our model under different settings as those in the 4D occupancy forecasting task.

We see that UniAD achieves the best overall performance, which exploits various types of auxiliary supervision to improve its planning quality.
Despite the strong performance, the additional annotations in the 3D space are very difficult to obtain, making it difficult to scale to large-scale driving data.
As an alternative, OccWorld demonstrates competitive performance by employing 3D occupancy as the scene representation which can be efficiently obtained by accumulating LiDAR scans~\cite{surroundocc}.

We observe that using ground-truth 3D occupancy as inputs, our OccWorld-O outperforms the previous perception-prediction-planning-based method OccNet~\cite{tong2023scene} by a large margin without using maps and bounding boxes as supervision, demonstrating the superiority of the world-model paradigm for autonomous driving.
Our end-to-end models OccWorld-D and OccWorld-T also demonstrate competitive performance using only 3D occupancy as supervision and OccWorld-S delivers non-trivial results with no supervision other than the future trajectory, showing the potential for interpretable end-to-end autonomous driving.

Though our model demonstrates very competitive L2 error, it slightly falls behind on the collision rate.
This is because it is more difficult to learn safe trajectories without the guidance of freespace or bounding box.
Still, OccWorld-O demonstrates comparable collision rates with OccNet which exploits map and box supervision, showing that OccWorld can learn the concept of freespace with 3D occupancy. 

We also observe that OccWorld shows excellent short-term planning performance (1s), but worsens quickly when planning longer futures.
For example, OccWorld-O achieves the best L2 error at 1s among all the methods but reaches 1.99 at 3s compared to 1.65 of UniAD.
This might result from the diverse future generations of world models, which might deviate from the ground-truth trajectory.

\textbf{Analysis of the scene tokenizer.}
We analyze the effect of different hyperparameters for the scene tokenizer in Table~\ref{tab:vae}.
The setting denotes latent spatial resolution, latent channel dimension, and the codebook size.
We see that using a larger codebook than 512 leads to overfitting and using a smaller codebook, spatial resolution, or channel dimension might not be enough to capture the scene distribution.
The reconstruction accuracy greatly improves with a larger spatial resolution, yet leads to poor forecasting and planning performance.
This is because the tokens cannot learn high-level concepts and are difficult to forecast the future.

\textbf{Analysis of the spatial-temporal generative transformer.}
We conducted an ablation study on both 4D occupancy forecasting and motion planning to analyze the design of the proposed spatial-temporal generative transformer, as shown in Table~\ref{tab:world}.
w/o spatial attn denotes discarding spatial aggregation and directly applying temporal attention to the input tokens.
w/o temporal attn represents that we replace the temporal attention with a simple convolution to output the next scene using the current world tokens.
w/o ego represents that we discard the ego token.
w/o ego temporal represents that we replace the temporal attention of the ego token with a simple MLP.
We observe that using spatial aggregation to model spatial dependencies and using temporal attention to integrate history information is vital to the performance of both 4D occupancy forecasting and motion planning tasks.
Also, only performing the 4D occupancy forecasting task without predicting motion reduces the performance. 
This verifies the effectiveness of joint modeling of scene evolutions and ego trajectories.
Finally, discarding the ego temporal attention leads to poor planning and surprisingly worse 3D forecast occupancy performance.
We think this is because integrating a wrongly predicted ego trajectory will mislead the forecasting.

%% file: sec/5_conclusion.tex
\section{Conclusion}
In this paper, we have presented a 3D occupancy world model (OccWorld) to model the joint evolutions of ego movements and surrounding scenes.
We have employed a 3D occupancy scene tokenizer to extract high-level concepts and used a spatial-temporal generative transformer for future prediction in an auto-regressive manner.
Both quantitive and visualization results have shown that OccWorld can effectively predict future scene evolutions in the comprehensive 3D semantic occupancy space.
We believe that OccWorld has paved the way for interpretable end-to-end autonomous driving without additional supervision signals.